\def\year{2020}\relax
\documentclass[letterpaper]{article} % DO NOT CHANGE THIS
\usepackage{aaai20}  % DO NOT CHANGE THIS
\usepackage{times}  % DO NOT CHANGE THIS
\usepackage{helvet} % DO NOT CHANGE THIS
\usepackage{courier}  % DO NOT CHANGE THIS
\usepackage[hyphens]{url}  % DO NOT CHANGE THIS
\usepackage{graphicx} % DO NOT CHANGE THIS
\usepackage{graphicx}  % DO NOT CHANGE THIS
\usepackage{amsmath, amssymb}
\usepackage{amsfonts}
\usepackage{multirow}
\usepackage{booktabs}
\usepackage{array}
\usepackage{verbatim}
\usepackage{color}
\usepackage{makecell}
\usepackage{bm}
\newcommand{\etal}{{\em et al\,.}}       % et al.
\newcommand{\eg}{{\em e.g.}}           % e.g.
\newcommand{\ie}{{\em i.e.}}           % i.e.
\title{Differentiable Meta-learning Model for Few-shot Semantic Segmentation}
\author{ Pinzhuo Tian\textsuperscript{\rm 1}, Zhangkai Wu\textsuperscript{\rm 1}, Lei Qi\textsuperscript{\rm 1}, Lei Wang\textsuperscript{\rm 2},
Yinghuan Shi\textsuperscript{\rm 1}\thanks{Corresponding author}, Yang Gao\textsuperscript{\rm 1} \\
\textsuperscript{\rm 1}National Key Laboratory for Novel Software Technology, Nanjing University, China\\
\textsuperscript{\rm 2}School of Computing and Information Technology, University of Wollongong, Australia \\ %If you have multiple authors and multiple affiliations
\{tianpinzhuo, qilei.cs\}@gmail.com wuzhangkai@smail.nju.edu.cn leiw@uow.edu.au \{syh, gaoy\}@nju.edu.cn% email address must be in roman text type, not monospace or sans serif
}

\begin{document}

\maketitle

\begin{abstract}
To address the annotation scarcity issue in some cases of semantic segmentation, there have been a few attempts to develop the segmentation model in the few-shot learning paradigm.
However, most existing methods only focus on the traditional $1$-way segmentation setting (\ie, one image only contains a single object). This is far away from practical semantic segmentation tasks where the $K$-way setting ($K\!>\!1$) is usually required by performing the accurate multi-object segmentation.
To deal with this issue, we formulate the few-shot semantic segmentation task as a learning-based pixel classification problem, and propose a novel framework called MetaSegNet based on meta-learning. In MetaSegNet, an architecture of embedding module consisting of the global and local feature branches is developed to extract the appropriate meta-knowledge for the few-shot segmentation. Moreover, we incorporate a linear model into MetaSegNet as a base learner to directly predict the label of each pixel for the multi-object segmentation. Furthermore, our MetaSegNet can be trained by the episodic training mechanism in an end-to-end manner from \emph{scratch}. Experiments on two popular semantic segmentation datasets, \ie, PASCAL VOC and COCO, reveal the effectiveness of the proposed MetaSegNet in the $K$-way few-shot semantic segmentation task.
\end{abstract}

\section{Introduction}
\noindent Recently, deep learning has made significant breakthroughs in many applications (\eg, image classification~\cite{DBLP:conf/cvpr/HeZRS16}, object detection~\cite{DBLP:conf/iccv/Girshick15} and semantic segmentation~\cite{DBLP:conf/cvpr/LongSD15}). However, its main challenge is that a large amount of labeled data is usually required to train deep models, which is impractical in real-world applications. Meanwhile, the accurate labeling is extremely laborious and expensive, particularly for pixel-wise annotation (\eg, semantic segmentation) and 3-D delineation (\eg, medical tissue labeling). Therefore, few-shot learning has recently drawn an increasing interest in the machine learning community, which can aid deep models to effectively learn knowledge from a few samples.
\begin{figure}
  \centering
  \includegraphics[width=8cm]{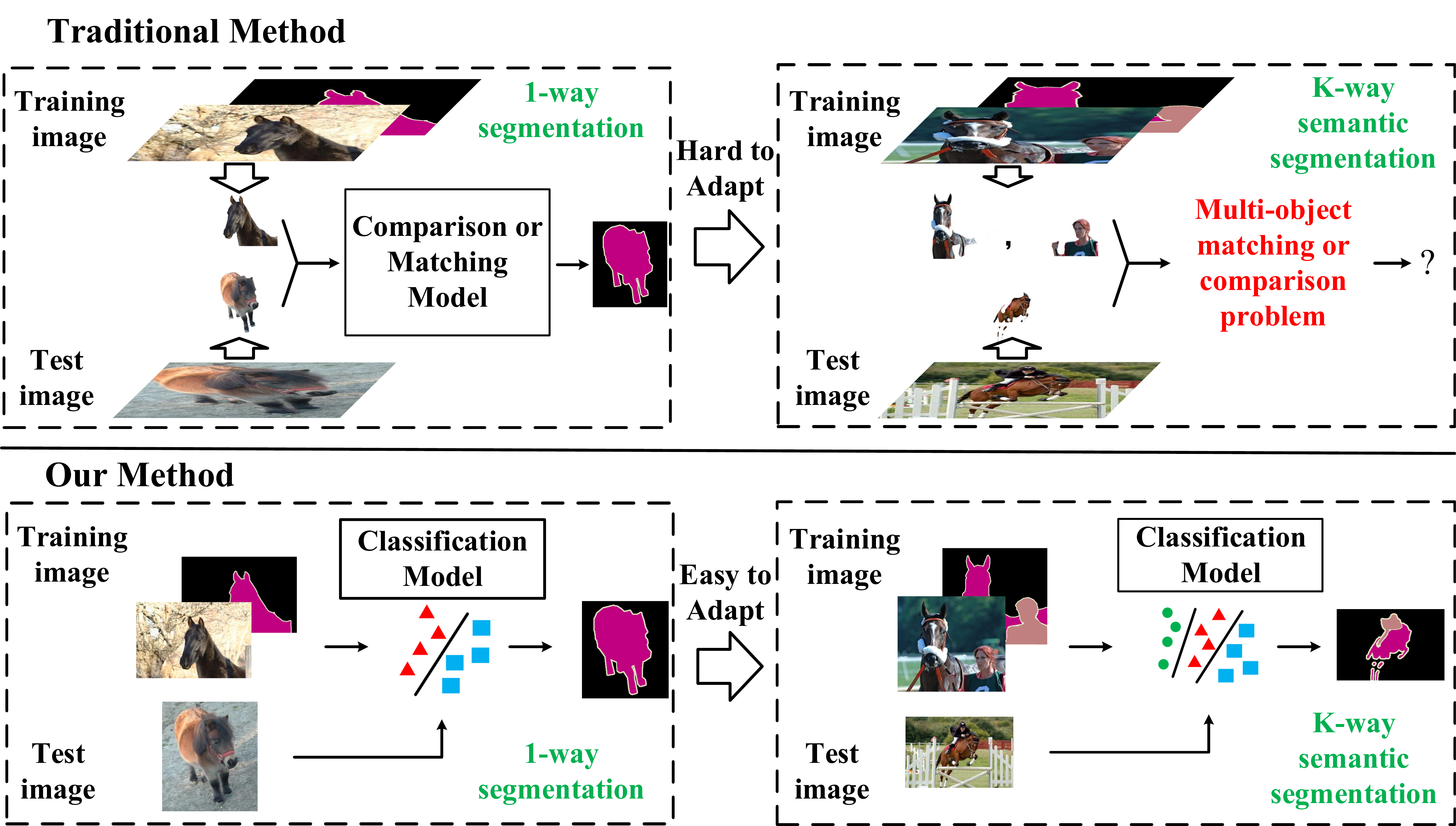}
  \caption{Comparison between our framework and traditional methods. The traditional methods use comparison or matching model in the $1$-way segmentation task to measure the similarity between training and test image, which is hard to extend to $K$-way semantic segmentation task. Differently, our method leverages learning-based model to classify the pixels in the test image for $K$-way semantic  segmentation.}
  \label{fig:introduction}
\end{figure}

Formally, few-shot learning aims to achieve good performance on the new classes with only a few labeled training data. Recently, most existing methods employ the meta-learning algorithm to deal with the few-shot learning problem \cite{DBLP:conf/nips/SnellSZ17,DBLP:conf/icml/FinnAL17}, and obtain incredible success in the classification task. However, there are only a few methods that attempt to develop few-shot learning algorithms for a more difficult task (\eg, semantic segmentation).

Most existing methods for few-shot semantic segmentation usually introduce the extra knowledge to acquire abundant prior knowledge to help the few-shot learning. However, they still suffer from several limitations as:

\begin{enumerate}
    \item \textbf{Distribution divergence}. Most existing methods require their model to be pre-trained on ImageNet \cite{DBLP:conf/cvpr/DengDSLL009} as a prerequisite. However, the distribution of these images in the few-shot segmentation task might be very different from ImageNet.

    \item \textbf{Hard to extend}. The existing methods usually solve the few-shot segmentation by utilizing metric-learning methods to measure the one-to-one similarity between training and test images in the $1$-way setting. However, these methods cannot be directly adapted to measure multi-object similarity in the $K$-way setting. %Thus most of previous methods aim to deal with $1$-way segmentation task that is far away from practical semantic segmentation setting.

    \item \textbf{Difficult to train}. These existing methods usually employ complex embedding models to learn feature representation for image comparison, which are hard to optimize. Thus, if we directly train them from scratch, they cannot work effectively in the target task.

\end{enumerate}

To address the aforementioned limitations, we propose a novel framework for few-shot semantic segmentation termed as MetaSegNet. Different from existing works, we formulate the few-shot segmentation as a dense prediction task, which classifies each pixel and is readily extended to the K-way setting ($K\!>\!1$). Specifically, we integrate a linear classification model into our framework as classifier, which could guarantee that our framework can be effectively optimized by the episodic training mechanism with meta-learning paradigm. Furthermore, a novel embedding module is designed in our framework to combine the global context and local information for better segmentation. Compared with two separate embedding modules in existing works, our embedding module is simple but effective. Meanwhile, the linear classification model combined with the new embedding module allows our framework to be effectively trained by an end-to-end manner from scratch for the multi-object semantic segmentation.

We evaluate our framework on two semantic segmentation datasets, \ie, PASCAL VOC 2012 \cite{DBLP:conf/cvpr/HariharanAGM15} and COCO 2014 \cite{DBLP:conf/eccv/LinMBHPRDZ14}, in the $K$-way, $N$-shot semantic segmentation task. Thanks to the effectiveness and efficiency of linear classifier and embedding module, our MetaSegNet can achieve the state-of-the-art performance in $K$-way, $N$-shot semantic segmentation task by using extremely small deep network without any extra knowledge.
Our main contributions are summarized as follows.

\begin{itemize}
  \item We formulate the few-shot semantic segmentation from the classification perspective, and propose a novel framework for dealing with the $K$-way, $N$-shot semantic segmentation problem.

  \item As far as we know, we are the first one to leverage the linear classifier instead of nonlinear layer to effectively train deep models in few-shot semantic segmentation.

  \item A simple embedding module with novel architecture is proposed, which can extract global and local information for semantic segmentation.

  \item The proposed novel framework is trained in an end-to-end manner from scratch. Also, it does not need any prior knowledge, which is the merit for the few-shot learning.

\end{itemize}

\begin{figure*}
  \centering
  \includegraphics[width=17cm]{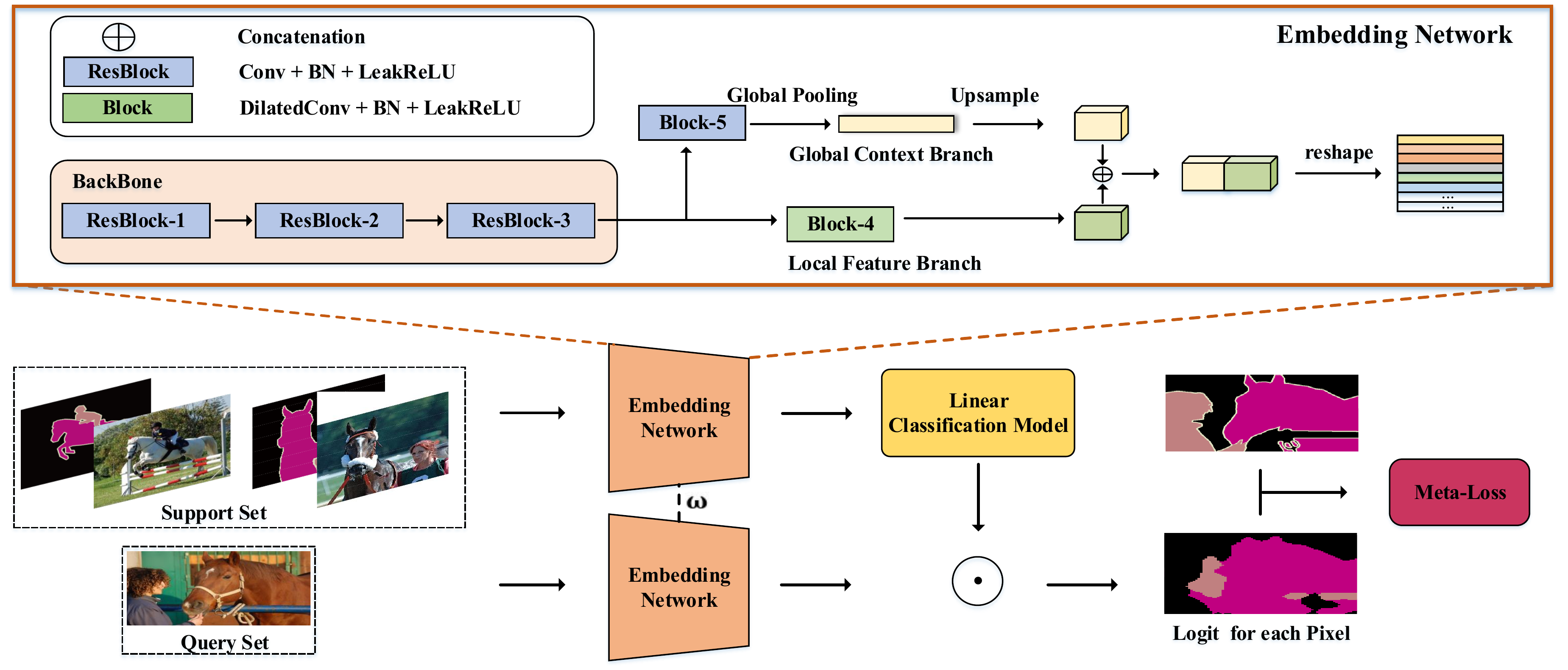}

  \caption{Pipeline of MetaSegNet for $2$-way semantic segmentation. In this framework, the support set is used to train a linear classification model, and we make use of meta-loss to train an embedding model, which can generalize well across tasks. Particularly, we put forward a novel embedding model, where two branches are utilized to extract local and global features for the pixel-level classification.}\label{fig:framework}

\end{figure*}

\section{Related Work}
In this part, we review the related works inclduing meta-learning and few-shot segmentation.

\indent \textbf{Meta-learning for few-shot Learning}.
Meta-learning studies what aspects of the learner~(commonly referred to bias or prior) effect generalization across a distribution of tasks ~\cite{DBLP:journals/air/VilaltaD02}. Nowadays, meta-learning approaches for few-shot learning can be broadly divided into three categories. Basically, (1) the \emph{metric-based methods} learn a sophisticated comparison model conditioned on distance or metric between training and test data. For example, the matching networks~\cite{DBLP:conf/nips/VinyalsBLKW16} develop an end-to-end differentiable nearest neighbor to perform comparison. Snell \etal~\cite{DBLP:conf/nips/SnellSZ17} propose a prototypical network which represents each category by the mean embedding of the examples, and utilize cosine distance to measure the similarity between test images and the prototypes. Li \etal~\cite{DBLP:conf/aaai/LiXHWGL19} design a covariance metric network for distance metric. (2) The \emph{model-based methods} wish to learn a parameterized predictor to estimate model parameters for new tasks. For example, Ravi \etal~\cite{DBLP:conf/iclr/RaviL17} utilize LSTM~\cite{DBLP:journals/neco/HochreiterS97} to predict parameters analogous to a few steps of gradient descent in parameter space. (3) The \emph{optimization-based methods} employ the optimization-based methods (\eg, gradient descent) to directly adapt the parameters to study knowledge across task. For example, Finn \etal~\cite{DBLP:conf/icml/FinnAL17} propose MAML algorithm aiming to learn a good initialization in a new task. Franceschi \etal~\cite{DBLP:conf/icml/FranceschiFSGP18} use bi-level optimization method to optimize meta-learning algorithm.

\indent \textbf{Few-shot segmentation}.
Previous methods for few-shot segmentation generally use the metric-based meta-learning framework, meanwhile two-branch separate embedding modules pre-trained on the ImageNet are used to compare the similarity between training and test images. These methods can be divided into two categories, according to the task setting.
(1) $1$-way segmentation. Shaban \etal~\cite{DBLP:conf/bmvc/ShabanBLEB17} first consider the few-shot learning problem in segmentation, and propose the two-branch comparison structure. Hu \etal ~\cite{DBLP:conf/aaai/HuYZYMS19} use attention and multi-context to enhance the two-branch comparison model. Also, Zhang \etal~\cite{DBLP:journals/corr/abs-1903-02351} introduce a two-branch class-agnostic segmentation network which contains a dense comparison module and an iterative optimization module. (2) $K$-way semantic segmentation. Dong \etal~\cite{DBLP:conf/bmvc/DongX18} and Wang \etal~\cite{wang2019panet} employ prototypes to represent the different semantic categories in the $K$-way semantic segmentation. However, they leverage different modules to align and measure the similarity between different objects.

Differently, our approach employs the optimization based meta-learning for few-shot segmentation. The linear classifier is integrated into our framework to deal with pixel classification. Different from $1$-way segmentation methods, our framework is easily extended to $K$-way semantic segmentation setting. Moreover, compared with~\cite{DBLP:conf/bmvc/DongX18,wang2019panet}, our framework is much simpler so that it could be optimized by an end-to-end manner without auxiliary information or data.

\section{Our Method}

In this section, we first introduce the $K$-way, $N$-shot segmentation task. Then we formulate the meta-learning framework. Finally, we elaborate each component in our method.

\subsection{$K$-way, $N$-shot Few-shot Semantic Segmentation}
In this paper, we consider the $K$-way, $N$-shot segmentation task which is seldom investigated ($K\!>\!1$). Here, $K$ denotes the number of classes, and $N$ is the number of training examples per class. Especially, in the few-shot learning setting, $N$ is usually small, \eg, $N \in \{1,5\}$. The training task $\mathcal{T}_i = (D^{\text{train}}_i, D^{\text{test}}_i) $ in meta-training set is sampled from a large-scale annotated dataset that contains multiple different training categories (\eg, we use $C^{\text{train}}$ to represent the set of training classes).
The training categories should be non-overlapping with the new classes (\eg, $C^{\text{novel}}$) in meta-testing set, \ie, $C^{\text{train}} \cap C^{\text{novel}} = \varnothing$. Firstly, the $K$ categories $C_i = \{C_i^k\}_{k=i}^K$ in $\mathcal{D}^{\text{train}}_i$
 are randomly selected from $C^{\text{train}}$. Then, $N$ images per category are sampled to constitute the whole support (training) set $\mathcal{D}^{\text{train}}_i$~(\ie, $\mathcal{D}^{\text{train}}_i = \{(x_n,y_n)~|~ n=1,\ldots,N\times K, y_n \in C_i\}$. Finally, the query~(test) set $D^{\text{test}}_i = \{ (x_n, y_n)~|~n = 1,\ldots, Q\times K, y_n \in C_i \} $ consists of $Q$ images per category.

However,
%different from the few-shot classification problem, the image in semantic segmentation task may contain multiple semantic class objects. Thus, we sample $N$ images for each category which contain this semantic object, and the semantic objects beyond the $K$ categories in the images are as background.
most exiting few-shot segmentation methods follow the paradigm and notations in~\cite{DBLP:conf/bmvc/ShabanBLEB17}, which is a special case of $K$-way, $N$-shot semantic segmentation, \ie, $K = 1$. Although the model can forward $K$ times to finish $K$-way semantic segmentation task, these models will suffer from the multi-object alignment problem. For example, one object may be classified to different semantic categories during $K$-time running, which obstructs traditional $1$-way segmentation model achieving good performance in the $K$-way setting.

 %Because, in each forward, a chosen semantic class is as foreground, the other semantic objects which may belong to the $K$ classes are regarded as background. The models for $1$-shot, $N$-way segmentation task just segment the foreground class, and don't need to recognize which object appear in the foreground.

%The definition of $K$-way, $N$-shot few-shot semantic segmentation in this  paper is different with~\cite{DBLP:conf/bmvc/DongX18}, which decomposed the multiple classes annotation of an image into multiple single-class annotations~(\eg, binary mask). Concretely, for each category, this semantic category object is regarded foreground, while, all other regions are viewed as background. Differently, in our definition, we allow multi-category annotation in  mask, which is consistent with the traditional semantic segmentation task.
%The setting of $K$-way, $N$-shot semantic segmentation problem in our paper is more similar with ordinary semantic segmentation problem than~\cite{DBLP:conf/bmvc/DongX18}, and is more difficult than $1$-way,$N$-shot segmentation task.

\subsection{Problem Formulation for Meta-learning}

Formally, given $M$ tasks, we denote the training task set~(\ie, meta-training set) as $\mathcal{T}=\{ ( \mathcal{D}^{\text{train}}_i, \mathcal{D}^{\text{test}}_i ) \}^{M}_{i=1}$, where the tuple $(\mathcal{D}^{\text{train}}_i, \mathcal{D}^{\text{test}}_i)$ represents the training data (\ie, support set) and test data (\ie, query set) for the $i$-th task $\mathcal{T}_i$. The goal of meta-learning is to enable a learning algorithm to effectively adapt to new task by generalizing from $\mathcal{T}$.

In particular, meta-learning approaches for few-shot learning usually introduce a generic embedding model $f_{\phi}$ parameterized by $\phi$ as a meta learner to map the task-specific domain into a common feature space. Also, we denote a base learner as ${\Lambda}$. It is a kind of learning model, which works in the individual task to estimate parameter $\mathbf{w}_i$ for the task $\mathcal{T}_i$. Specifically, this process can be formulated as:

\begin{equation}\label{eq:base-learner}
   \Lambda(f_{\phi}(\mathcal{D}^{\text{train}}_i))
       = \min_{\mathbf{w}_i} \mathbb{E}_{(x_t, y_t) \in \mathcal{D}^{\text{train}}_i} [\mathcal{L}^{\text{base}} (f_{\phi} (x_t), y_t; \mathbf{w}_i)],
\end{equation}
where $\mathcal{L}^{\text{base}}$ represents the loss function of the base learner. $x_t$, $y_t$ denote the training sample and its label in support set $\mathcal{D}_i^\text{train}$ of task $\mathcal{T}_i$, respectively. The squared loss is used as $\mathcal{L}^{\text{base}}$ in this paper.

However, if we ignore the knowledge that can be transferred between tasks and only re-train a predictor for new task with a few examples, the predictor cannot achieve satisfactory performance. For this issue, the meta learner aims to minimize the generalization~(or test) error across tasks, and helps the base learner to improve performance. The objective used to calculate the parameter $\phi$ of the meta learner $f_{\phi}$ can be written as:

\begin{equation}\label{eq:meta-learner}
\begin{split}
  & \min_{\phi}  \sum_{i=1}^{M} \mathcal{L}^{\text{meta}} (\mathcal{D}^{\text{test}}_i;\mathbf{w}_i,\phi) \\
   = & \min_{\phi}  \sum_{i=1}^{M} \mathcal{L}^{\text{meta}} (\mathcal{D}^{\text{test}}_i; \Lambda(\mathcal{D}^{\text{train}}_i; \phi) ,\phi),
\end{split}
\end{equation}
where $\mathcal{L}^{\text{meta}}$ denotes meta-learning loss function.
In this paper, we adopt the cross-entropy loss for $\mathcal{L}^{\text{meta}}$.

In fact, CNN is usually used as the embedding model $f_{\phi}$. In order to optimize the $\phi$ in Equation (\ref{eq:meta-learner}) by back-propagation and stochastic gradient descent~(SGD), $\Lambda$ is preferred to be a simple and efficient model (\eg, nearest neighbor methods or linear models)~\cite{DBLP:conf/iclr/BertinettoHTV19}. Once the embedding model $f_{\phi}$ is learned, the base learner for a new task $\mathcal{T}_{\text{new}} = ( \mathcal{D}^{\text{train}}_{\text{new}}, \mathcal{D}^{\text{test}}_{\text{new}})$ can be got by Equation (\ref{eq:base-learner}).

\subsection{The Proposed MetaSegNet}
In this section, we describe each module of the proposed MetaSegNet in detail.
\subsubsection{Embedding network.}
In order to effectively extract the meta-knowledge for $K$-way, $N$-shot semantic segmentation, we propose a novel architecture of embedding network as a meta learner in our framework as illustrated in Figure~\ref{fig:framework}.

The proposed embedding network consists of two sub-modules, \ie, the feature extractor and the feature fusion module. The architecture of our embedding network is different from the existing methods which utilize two separate modules to extract features of support and query set, respectively \cite{DBLP:conf/bmvc/ShabanBLEB17,DBLP:conf/aaai/HuYZYMS19}. In our framework, we use one feature extractor to extract the local and global feature simultaneously for support and query set, and the fusion module is used to fuse local and global information for better predicting the label of each pixel.

Besides, most existing methods directly use VGG-16 or ResNet-50 pre-trained on the ImageNet as the backbone. To tackle the distribution divergence as aforementioned, we train MetaSegNet from scratch without introducing any extra information. So, we just use ResNet-9 which has fewer parameters as backbone in order to prevent the possible over-fitting. With the same goal of these previous semantic segmentation methods, our framework is also required to conduct the dense prediction for a full-resolution output. Specifically, to handle the conflicting demands of multi-scale reasoning and full-resolution dense prediction, we first remove the max-pooling layers behind block-3 and block-4, and use the dilated  convolution~\cite{DBLP:journals/corr/YuK15} instead of the traditional convolution in block-4 to aggregate the multi-scale contextual information.

In particular, we argue that the dilated convolution in block-4 is not enough to extract the local information for semantic segmentation. Inspired by~\cite{DBLP:journals/corr/LiuRB15}, we add an additional branch to extract the global context to help semantic segmentation. Typically, two max-pooling layers are inserted ahead and behind the block-5 in the global context branch to expand the receptive fields.

In the feature fusion module, we first unpool~(replicate)~global feature to the same size of local feature map spatially and then concatenate them together. Since the two kinds of features could have different scales and norms, we apply $\ell_2$-norm in each channel for normalization. After this normalization, we reshape the combined feature map to pixel feature map for pixel-wise classification. %That is to say, if $Z \in \mathbb{R} ^{h\times w \times c}$ is a combined feature map for an image from feature extractor module, $Z$ will be reshaped to $P \in \mathbb{R}^{n \times c}$, where $n=h \times w$ indicates the pixel feature map for the pixels in the image.

\begin{table}[]
\centering
\footnotesize{
\begin{tabular}{cccc|c}
%{p{50 pt}<{\centering} p{40 pt}<{\centering}  p{25 pt}<{\centering} c  p{15 pt}<{\centering}   p{15 pt}<{\centering}  p{15 pt}<{\centering}  p{15 pt}<{\centering}  p{25 pt}<{\centering}}
\toprule
\textbf{Model}& \textbf{Params(M)} & \makecell{\textbf{Without} \\ \textbf{pre-} \\ \textbf{training}} & \makecell{\textbf{End-}\\ \textbf{to-end}}  &
\makecell{\textbf{Mean} \\ \textbf{IoU(\%)} } \\
\toprule
 Fine-tuning & 14.2   & $\bm{\times}$ & \checkmark & 28.6 \\
  Prototype & 14.2 & $\bm{\times}$ & \checkmark  & 35.0    \\

  SG-One & 19 & $\bm{\times}$ & \checkmark & 39.4 \\
  PANet & 14.2 & \checkmark & \checkmark &41.3 \\

  PLSEG  & 28.4 & $\bm{\times}$ &  $\bm{\times}$ & 42.6          \\

\makecell{MetaSegNet \\ (ours)} & \textbf{13.1} &  \checkmark  & \checkmark     &\textbf{43.3}    \\
\bottomrule
\end{tabular}
}

\caption{The performance of $2$-way, $5$-shot semantic segmentation on PASCAL-5$^2$. The proposed MetaSegNet needs no extra knowledge. Moreover, the parameter of MetaSegNet is least among all models.}

\label{Tab:mulitresultpascal}
\end{table}

\subsubsection{Differentiable linear base learner.}
According to \cite{DBLP:conf/icml/FranceschiFSGP18,DBLP:conf/icml/BalcanKT19}, the choice of base learner is important for optimization based meta-learning methods. Because the remaining embedding model is optimized to generalize across different tasks in Equation (\ref{eq:meta-learner}), the parameter of base learner $\Lambda$ needs to be calculated in an efficient way. Moreover, when we apply the stochastic gradient descent~(SGD) to estimate the parameters $\phi$ of the embedding model, the gradients of the test error $\mathcal{L}^{\text{meta}}(\mathcal{D}^{\text{test}}; \mathbf{w}, \phi)$ with respect to $\phi$ should be precisely computed.

The no-parametric metric based model avoids the optimizing parameter in base learner. And the linear model is easy to compute the derivative of the linear model parameter $\mathbf{w}$ with respect to $\phi$.
Their efficacy has been proven in few-shot classification problem \cite{DBLP:journals/corr/abs-1904-03758}. However, the classical semantic segmentation frameworks~\cite{DBLP:conf/cvpr/LongSD15,DBLP:journals/pami/ChenPKMY18} use deconvolution layer as segmentation head to predict label for each pixel. It is difficult to calculate the analytic or optimal solution for deconvolution layer, thus if we directly use it as base learner in  Equation (\ref{eq:meta-learner}), the whole framework will be hard to train. Therefore, we choose the differentiable linear model (\ie, ridge regression) as base learner in our framework. Hence, the base learner $\Lambda$ can be written as:

\begin{equation}\label{eq:linear-model}
    \Lambda(\mathbf{X}) =  \arg \min_{\mathbf{w}} \|\mathbf{X}\mathbf{w} - \mathbf{y}\|^2 + \lambda\|\mathbf{w}\|^2,
\end{equation}
where $\mathbf{X} \in \mathbb{R}^{n \times c}$ is the pixel feature matrix obtained from the embedding network. $ \mathbf{y} \in \mathbb{R}^{n}$ is the label of each pixel.

Furthermore, there is a closed form solution for Equation (\ref{eq:linear-model}). Support for task $\mathcal{T}_i = ( D^{\text{train}}_i, D^{\text{test}}_i)$, we can directly obtain $\mathbf{w}$ as:

\begin{equation}\label{eq:closedform}
  \mathbf{w} =  (\mathbf{X}^\top \mathbf{X} + \lambda I)^{-1} \mathbf{X}^\top \mathbf{y}.
\end{equation}

Due to the property of Equation (\ref{eq:linear-model}) (\ie, there exists the closed solution), it is very easy to integrate ridge regression into meta-learning optimization framework. We can directly optimize parameter $\phi$ of the embedding model in Equation (\ref{eq:meta-learner}) by back-propagation algorithm.

In fact, the ridge regression was originally designed for the regression task, we also adjust the prediction of base linear $\Lambda$ by Equation  (\ref{eq:adjust}), as in~\cite{DBLP:conf/iclr/BertinettoHTV19}.

\begin{equation}\label{eq:adjust}
  \hat{\mathbf{y}} = \alpha \mathbf{X}' \mathbf{w} + \beta,
\end{equation}
where $\mathbf{X}' \in \mathbb{R}^{n \times c}$ is the feature matrix of the test image. The $\alpha \in \mathbb{R}$ and $\beta \in \mathbb{R}$ denote the scale and bias, respectively.

\subsubsection{Meta-learning objective.}
In the proposed framework, we adopt the cross-entropy loss to evaluate the pixel-wise prediction. And we regard $\lambda$ in Equation (\ref{eq:closedform}) and $\alpha$, $\beta$ in Equation (\ref{eq:adjust}) as learnable parameters. Finally,
the meta-learning objective can be written as:

\begin{equation}\label{eq:meta-objective}
  \begin{split}
     &\mathcal{L}^{\text{meta}} (\mathcal{T}; \mathbf{w}, \phi, \alpha, \beta, \lambda) = \\
     & - \frac{1}{|\mathcal{T}|\cdot|\mathcal{D}^{\text{test}}_{i}||\mathcal{I}|} \sum_{i \in \mathcal{T}} \sum_{(\mathbf{x},\mathbf{y}) \in \mathcal{D}^{\text{test}}_i} \sum_{p \in \mathcal{I}} \log (S_{p y_p}).
  \end{split}
\end{equation}
$\mathcal{I} $ is the set of pixels in the image. Let $s_{pc}$ be the score of pixel $p$ and class $c$, which can be got by Equation (\ref{eq:adjust}). $S_{pc} = \exp(s_{pc}) / \sum_{k=1}^{N} \exp(s_{pk})$ represents the softmax probability of class $c$ at pixel $p$.

\section{Experiments}
\subsection{Dataset, Metric and Implementation}
\textbf{Dataset}.
We perform the extensive evaluation on two semantic segmentation benchmark datasets, \ie,  PASCAL VOC and COCO: (1)
PASCAL VOC 2012 \cite{DBLP:conf/cvpr/HariharanAGM15} contains 20 different object categories which consists of 10,582 and 1,449 images as the training and validation sets, respectively.
%We and Hariharan et al.~\cite{DBLP:conf/cvpr/HariharanAGM15} augmented the annotation set~(SDS) consisting 10,582, 1,449 images in training, validation sets.
(2) COCO 2014 \cite{DBLP:conf/eccv/LinMBHPRDZ14} dataset is a challenging large-scale dataset which contains 80 different object categories. In COCO, 82,783 and 40,504 images are used for training and validation, respectively.

\textbf{Evaluation metric}.
To compare the performance of the different models, the average of mean intersection-over-union (mIoU) for each task is introduced for evaluation.

\textbf{Implementation details}.
 Our proposed MetaSegNet involves five residual blocks. Each residual block consists of three modules, and each module includes a $3\times3$ convolution with $k$ filters, a batch normalization and a LeakyReLU~(0.1). The size of five residual blocks are set as 64, 128, 256, 512 and 512, respectively. We set block-1 to block-2 as the backbone which is followed by a $2\times 2$ max-pooling.

In the global feature branch, two $2\times2$ max-pooling layers are inserted before and after block-5, respectively. A global pooling is employed to extract the global feature. In the local feature branch, the dilated convolution is introduced here to extract the multi-scale local features. Specifically, in the $1$-way segmentation setting, the dilation for block-4 is set as 2, 4 and 8, respectively. While, in the $K$-way setting, the dilated convolution is used in block-3 with dilation as 1, 2 and 4, and the dilation for block-4 is set as 8, 16 and 32.

For optimization, we use Adam \cite{DBLP:journals/corr/KingmaB14} with learning rate as 0.001. For PASCAL, our model is meta-trained for 40 epochs, and each epoch consists of 1000 episodes. For COCO, our model is meta-trained for 80 epoches, and each epoch consists of 500 episodes. During meta-training, we adopt horizontal flip, and randomly rotate the image with 0, 90, 180 or 270 degree for data augmentation. We implement our method by PyTorch on two NVIDIA 2080 Ti GPUs with 12 GB memory.

\begin{table}[]
\centering
\footnotesize{
\begin{tabular}{cccc}
\toprule
\textbf{Model} & \textbf{Shot=1} & \textbf{Shot=5} & \textbf{Shot=10} \\
\toprule
Prototype & 11.0 & 12.8& 14.3 \\
Fine-turning & 29.3 & 28.2 & 29.1\\
MetaSegConv &26.9 &27.3 &26.8\\
MetaSegNet (ours) &\textbf{33.2} &\textbf{37.9}& \textbf{38.3} \\
\bottomrule
\end{tabular}
}

\caption{The performance of $2$-way semantic segmentation on COCO. All the models are not pre-trained.}
\label{Tab:cocoresult}
\end{table}

\subsection{Results on PASCAL ($K$-way, $N$-shot)}
\textbf{Setting}.
PASCAL-5$^i$ \cite{DBLP:conf/bmvc/ShabanBLEB17} is a very popular few-shot segmentation benchmark, thus we also report our results on this dataset.
In fact, PASCAL-5$^i$ comes from PASCAL, the 20 object categories in PASCAL are divided into 4 splits with three splits for training and the rest one for testing. Typically, because of the unbalance of the multi-object image in each split, we choose PASCAL-5$^2$ as the test split, and evaluate our method in $2$-way semantic segmentation setting, which are the same with \cite{DBLP:conf/bmvc/DongX18}. The images that only contain one object in training splits are used to complement the meta-training set. The images in PASCAL-5$^2$ which contain \emph{person} and another held-out class are sampled as support and query set for new tasks. In fact, there are four situations of the new tasks, \ie, \{(table, person),~(dog, person),~(horse, person),~(motobike, person)\}. We randomly sample 100 tasks for each situation, and calculate the average performance of the four situations. Query set in each new task has two images.

To evaluate our MetaSegNet in $2$-way semantic segmentation task on PASCAL-5$^2$, we compare it with two baseline models (\ie, baseline fine-tuning and baseline prototype model \cite{DBLP:conf/bmvc/DongX18} ), one $1$-way segmentation model SG-One (running 2 times) \cite{DBLP:journals/corr/abs-1810-09091} and two state-of-the-art $K$-way, $N$-shot semantic segmentation models (\ie, PLSEG \cite{DBLP:conf/bmvc/DongX18} and PANet \cite{wang2019panet}). The fine-tuning model is a standard FCN that employs a VGG-16 pretrained on the ImageNet as the backbone. The images from training classes are first used to train this backbone, then the support set in new task is used to fine-tune this model.

\textbf{Results}.
We report the results in Table \ref{Tab:mulitresultpascal}. In the $2$-way, $5$-shot setting, our framework achieves the state-of-the-art performance in spite of simple embedding model and no extra information. Although the information from ImageNet provides extra knowledge for baselines and SG-One model, the multi-object metric or matching problem hinders these approaches from reaching satisfied results in $K$-way setting. Differently, the proposed MetaSegNet can handle multi-object matching problem from classification perspective, and achieve good performance in this setting.
Furthermore, compared with the other state-of-the-art $K$-way semantic segmentation models (\ie, PLSEG and PANet), MetaSegNet can learn more useful cross-task meta knowledge, which can effectively guide the semantic segmentation in query set.
%Though, the PLSEG achieve $41.9\%$ mIoU in $2$-way, $1$-shot. The performance of MetaSegNet is  $35.43\%$. We think that the information from ImageNet is very important in the situation extremely lack of training samples, so the pre-training may enhance the performance of PLSEG with high probability and it's unfair to compare our method with PLSEG directly. Even the simple prototype baseline without elaborate metric method could achieve $34.8\%$, which can further verify our guess above.

%In $2$-way, $1$-shot setting, PLSEG achieves the best performance. We think that the information from ImageNet may provide help in this extremely lack of training samples. Even the simple prototype baseline without elaborate metric method could achieve quite good performance. However, in $2$-way, $5$-shot setting, with the information increasing from support set, our framework achieve the state-of-the-art performance in spite of simple embedding model and no extra information. And when the shots increasing, our framework could utilized the information from support set better than PLSEG. We think the information from support set is more suitable for the semantic segmentation in query set. The proposed MetaSegNet can handle multi-object matching problem in previous work. Although, the information from ImageNet provide a high starting point for PLSEG, the multi-object metric or matching problem hinder the metric based approach to utilize the increasing knowledge from support set. \textcolor{red}{I think we may need to re-organize this paragraph}

\begin{table}[]
\centering
\footnotesize{
\begin{tabular}{ccc|c}
\toprule
\multirow{2}{*}{\makecell*[c]{\textbf{Model}}} & \multirow{2}{*}{\makecell*[c]{\textbf{Params (M)}}} &
\multirow{2}{*}{ \makecell{\textbf{Extra} \\ \textbf{images}} } & \multirow{2}{*} {\makecell{\textbf{Mean} \\ \textbf{ IoU(\%)}}}\\
&&&\\
\toprule
co-FCN (param.) & 34.2 & $1.2\times10^6$  & 58.3\\
OSLSM &276.7& $1.2\times10^6$ & \textbf{61.5} \\
MetaSegNet (ours)& \textbf{13.1} & \textbf{0} & 53.4\\
MetaSegNet (ours) & \textbf{13.1} &$6\times10^4$ & 59.5\\
\bottomrule
\end{tabular}
}
\caption{ The results of $1$-way, $5$-shot segmentation on PASCAL-5$^i$. All results are computed by taking the average of 4 splits on PASCAL-5$^i$. Extra images represent the number of images for per-training the model. Although, MetaSegNet uses $1/20$ extra images and $1/21$ model size compared with OSLSM, the Mean IoU of MetaSegNet just decreases $2\%$ ($61.5$ vs. $59.5$) accuracy.}
\label{Tab:1way}
\end{table}

\begin{figure*}
  \centering
  \includegraphics[width=17.5cm]{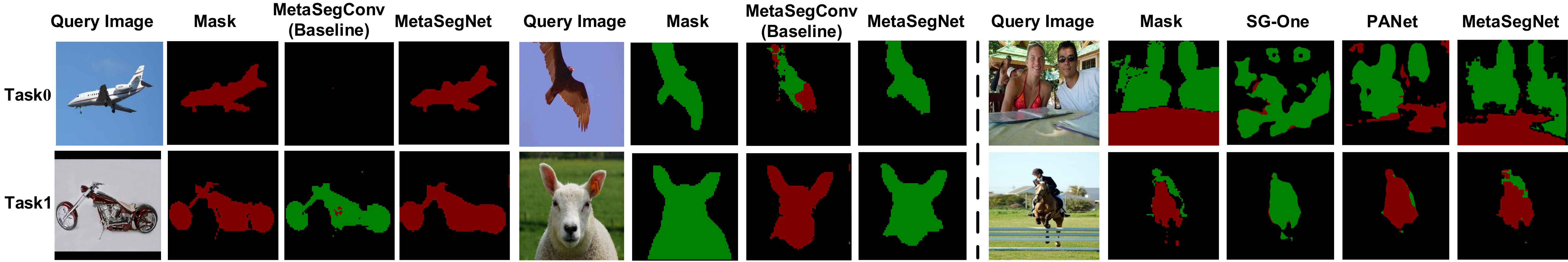}
  \caption{Visual examples for baselines and MetaSegNet. The left part is the results of two tasks in $2$-way, $5$-shot single object semantic segmentation on PASCAL-5$^{i}_{val}$. The right part is the results of two tasks in $2$-way, $5$-shot multi-object semantic segmentation on PASCAL-5$^2$.
  Red and green regions represent the labels of different semantic objects.}
  \label{fig:visualresult}
\end{figure*}

\subsection{Results on COCO ($K$-way, $N$-shot)}

\textbf{Setting}.
%COCO 2014 dataset is very larger and challenging. Directly experimenting on the original dataset is very demanding on time and computation.
Similar to \cite{DBLP:journals/corr/abs-1903-02351}, we choose a subset of the original COCO dataset to evaluate our model and baselines. We choose 18 categories as training classes, and 7 categories as new classes. The training class IDs are \{1, 2, 4, 5, 7, 8, 9, 19, 20, 21, 23, 24, 25, 62, 63, 64, 65, 70\}, and the testing class IDs are \{3, 6, 16, 17, 18, 22, 67\}. Because of the limitation of images that contain  multiple classes, we evaluate MetaSegNet on $2$-way with $1$-shot, $5$-shot and $10$-shot few-shot semantic segmentation tasks. Considering that the images in the training classes are more abundant than that in PASCAL, we directly sample images from training classes which contain 2 classes to complement the meta-training set. In addition, these images are cropped according to the bounding box of semantic objects to reduce the size of background. There are 1500 tasks in total sampled from new classes as meta-testing set, where the query set in each task includes 2 images.

According to our best knowledge, our work is the first attempt to deal with $K$-way, $N$-shot semantic segmentation on COCO 2014. For comparison, we design three baseline models. As the same as the previous experiment, the fine-tuning model is also a variant of traditional FCN. The embedding module in MetaSegNet without global context branch and $1\times 1$ conv layer are used as backbone and segmentation head for fine-tuning model, respectively. The images from training classes are used to train the model, and the support set for new task is used to fine-tune this model. The MetaSegConv utilizes the same model with fine-tuning model, however, MetaSegConv is trained by the optimization based meta-learning framework. The $1\times 1$ conv layer is regarded as a base learner that is optimized by one step by Adam with learning rate as 0.001. As for prototype baseline, the embedding module is the same as MetaSegNet. However, the ridge regression is replaced by Euclidean distance, and the mean of pixel features is as the prototype for one semantic category as in \cite{DBLP:conf/nips/SnellSZ17}.

\textbf{Results}.
The results are reported in Table \ref{Tab:cocoresult}. Although the COCO 2014 is more difficult than PASCAL as we know, our MetaSegNet still outperforms the baselines with a large margin in all cases. Also, we could notice that the performance of MetaSegNet increases with the increasing number of shot, while all the other baselines have no obvious variance.

\subsection{Extension to $1$-way, $N$-shot}
To fully investigate the performance of our MetaSegNet, we degenerate our proposed method from the $K$-way, $N$-shot task to the traditional $1$-way, $N$-shot task and report the performance on PASCAL-5$^i$.
%We also evaluate MetaSegNet in $1$-way, $N$-shot segmentation task on PASCAL-5$^i$, which is a specifical case of $K$-way, $N$-shot semantic segmentation.

\textbf{Setting}.
Since most of conventional methods are specifically designed to solve this $1$-way, $N$-shot task, for a fair comparison, we directly follow the same experiment settings as~\cite{DBLP:conf/bmvc/ShabanBLEB17}. In particular, one class is sampled from training classes as the foreground, and the other categories are regarded as the background. Thus, the mask in this case is a binary map. The four splits on PASCAL-5$^i$ are all used to evaluate MetaSegNet. In each split, 1000 tasks are sampled from new classes as the meta-testing set, and there is one image in query set for each new task.

Two state-of-the-art $1$-way segmentation models, \ie, co-FCN \cite{DBLP:conf/iclr/RakellySDEL18} and OSLSM \cite{DBLP:conf/bmvc/ShabanBLEB17}, are introduced for comparison. Since there is no multi-object metric case in this setting, the superiority of MetaSegNet is not so obvious when compared with these models. We also use extra images to pre-train our MetaSegNet. As using the entire ImageNet for pre-training is too heavy and time-consuming, we only pre-train our MetaSegNet on the miniImageNet (60,000 images) \cite{DBLP:conf/nips/VinyalsBLKW16}. Besides, we also evaluate MetaSegNet without pre-training.

\textbf{Results}.
We report the results in Table \ref{Tab:1way}. It is very convenient for metric-based methods to measure the similarity between foregrounds when each foreground only contains one object. Therefore, the metric-based approaches perform well in this setting. Although MetaSegNet is designed for $K$-way, $N$-shot task by using simple model and much less extra knowledge, it can still achieve a competitive result when compared with other 1-way methods in the 1-way setting.
%Similar with $K$-way, $N$-shot semantic segmentation task, more extra knowledge and complex embedding model provides a good start point for metric based method (see $1$-shot). However, these model can't learn how to effectively utilize the information from support set (see $5$-shot).

\subsection{Ablation Study}
Considering that there are no validation categories on PASCAL-5$^i$, we use a new few-shot semantic segmentation dataset, \ie, PASCAL-5$^i_{val}$, in ablation study. Similar to  PASCAL-5$^i$, the PASCAL VOC 2012 is also divided into 4 splits to evaluate our method. However, in each split, three classes are chosen as validation set (\ie, 6-8 in split 0; 11-13 in split 1; 16-18 in split 2 and 1-3 in split 3), and the other 12 classes are used as training classes. All images in new categories are sampled to make up support set and query set for new tasks in meta-testing set, just like miniImageNet~\cite{DBLP:conf/nips/VinyalsBLKW16} utilized for the few-shot classification task. Because the images containing multiple objects are limited on PASCAL-5$^i_{val}$, all the methods are evaluated in $2$-way single object semantic segmentation. Each image in query set contains one semantic object which belongs to the $K$ categories. We randomly sample 500 tasks as meta-testing set. There are four images in query set for new tasks~(\ie, two images for each class).

\begin{table}[]
\centering
\footnotesize{
\begin{tabular}{p{70 pt}<{\centering} p{30 pt}<{\centering} p{25 pt}<{\centering} p{26 pt}<{\centering} |c}
\toprule
\multirow{2}{*}{\textbf{Model}} & \multirow{2}{*}{ \makecell{\textbf{Episodic} \\ \textbf{Training}}} &
\multirow{2}{*}{ \makecell{\textbf{Linear} \\ \textbf{model}} }& \multirow{2}{*}{\makecell{\textbf{GC}\\\textbf{-branch}}} &\multirow{2}{*} {\makecell{\textbf{Mean} \\ \textbf{ IoU(\%)}}}\\
&&&&\\
\toprule
Fine-tuning&  & & & 27.28\\

MetaSegConv&  \checkmark& & & 35.00\\

MetaSegNet-NG& \checkmark& \checkmark & &37.75\\
MetaSegNet (ours)& \checkmark & \checkmark & \checkmark &\textbf{38.92}\\
\bottomrule
\end{tabular}
}
%\caption{ Ablation study on PASCAL-5$^1_{val}$ in $2$-way, $1$-shot setting. mIoU of the 4 splits are averaged as final results. GC-branch represents the global context branch in embedding model.}

\footnotesize{
\begin{tabular}{p{70 pt}<{\centering} p{30 pt}<{\centering} p{25 pt}<{\centering} p{26 pt}<{\centering} | c}
\toprule
\multirow{2}{*}{\textbf{Model}} & \multirow{2}{*}{ \makecell{\textbf{Episodic} \\ \textbf{Training}}} &
\multirow{2}{*}{ \makecell{\textbf{Linear} \\ \textbf{model}} }& \multirow{2}{*}{\makecell{\textbf{GC}\\\textbf{-branch}}} &\multirow{2}{*} {\makecell{\textbf{Mean} \\ \textbf{ IoU(\%)}}}\\

 & & & & \\
\toprule

Fine-tuning&  & & &31.95\\

MetaSegConv&  \checkmark& & & 36.04\\

MetaSegNet-NG& \checkmark& \checkmark & &48.34\\
MetaSegNet (ours)& \checkmark & \checkmark & \checkmark & \textbf{49.44}\\
\bottomrule
\end{tabular}
}
\caption{ Ablation study on PASCAL-5$^i_{val}$ in $2$-way setting. The top table is the results of $1$-shot, and the bottom table is the results of $5$-shot.
GC-branch represents the global context branch in embedding module.}
\label{Tab:ablationstudy}
\end{table}

\textbf{Global context branch}.
To verify efficacy of the global context branch, we evaluate the performance of MetaSegNet-NG which uses the embedding module without the global context branch. The dilation of block-4 is set as 2, 4, 8, which is the same with MetaSegNet. The experimental results are shown in Table \ref{Tab:ablationstudy}. As seen, the global context branch indeed benefits the few-shot semantic segmentation.

\textbf{Linear classification model}.
The MetaSegConv is used to evaluate the impact of the ridge regression, which employs the same embedding module with MetaSegNet-NG. Besides, MetaSegConv leverages $1\times1$ conv layer as base learner, unlike the ridge regression is utilized in MetaSegNet-NG. According to Table~\ref{Tab:ablationstudy}, the mIoU of MetaSegNet-NG outperforms MetaSegConv by $2.75\%$ ($37.75$ vs. $35.00$) and $12.3\%$ ($48.34$ vs. $36.04$) in $1$-shot and $5$-shot, respectively. The main reason is that the conv layer is too complex as base learner and it is not conductive to optimization whereas the ridge regression can avoid this issue.

Also, Table~\ref{Tab:ablationstudy} shows the performance of  fine-tuning in $K$-way, $N$-shot semantic segmentation to further confirm the power of meta-learning in few-shot semantic segmentation.

\begin{figure}[h]
  \centering
  \includegraphics[scale = 0.35]{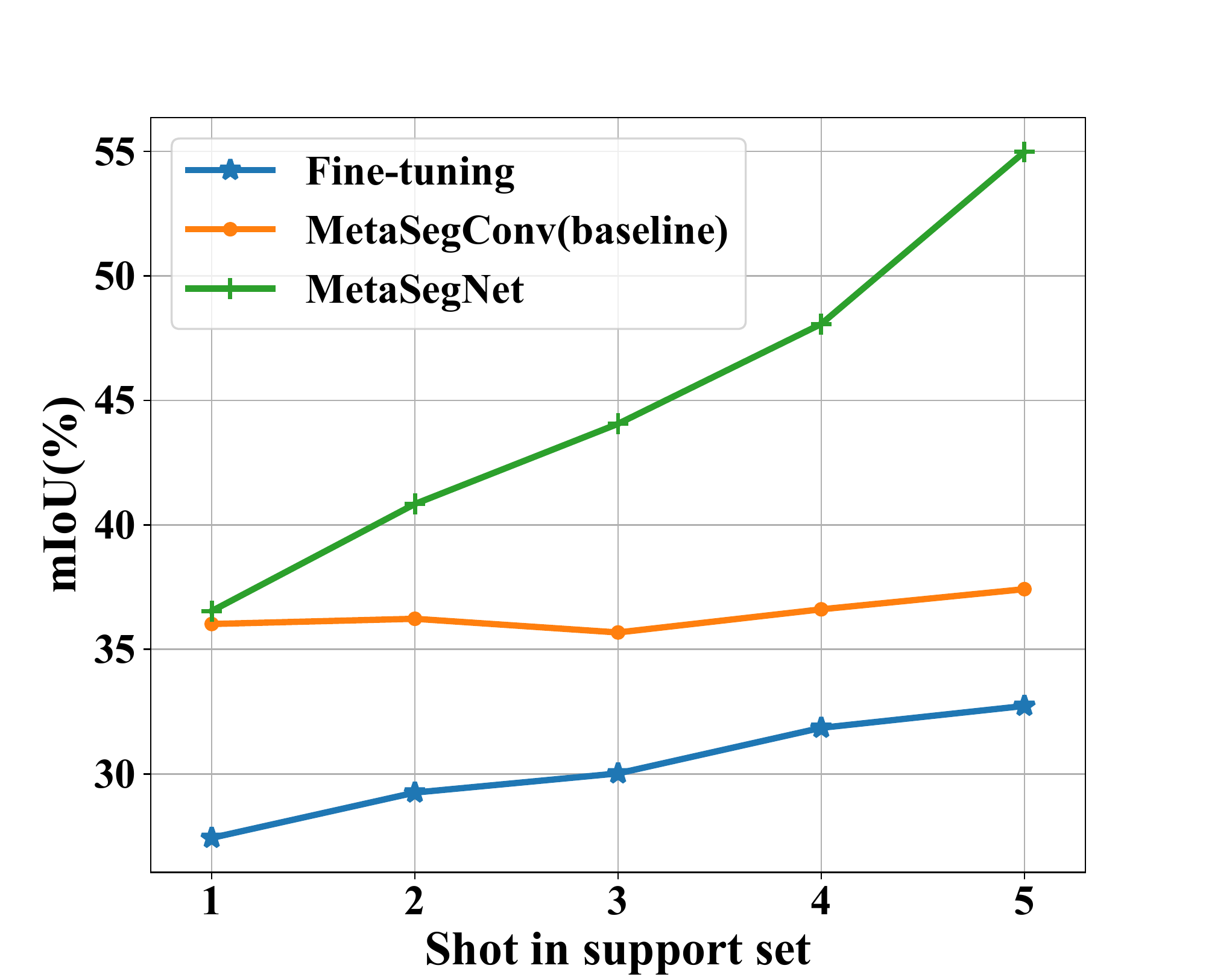}
   \caption{The results of different shots in support set for $2$-way semantic segmentation on PASCAL-5$^1_{val}$.}
  \label{fig:shots}
\end{figure}

\textbf{Efficacy of shot in support set.}
In this part, we research the impact of shot in support shot during meta-testing. As seen in Figure~\ref{fig:shots}, the performance of our method is in positive correlation with the shot in support set. However, the change of the baselines is not very obvious. This is attributable to our framework can learn how to efficiently use the information from support set by meta-learning framework. Just like human, we cannot learn well in extremely short of information (\eg, $1$-shot) in a new task, but our skill to solve a new task is in positive correlation with attempts.

\section{Conclusion}
In this paper, we aim to solve the $K$-way, $N$-shot few-shot semantic segmentation task, which is more difficult than the conventional $1$-way setting. Different from the existing works, we formulate this task as a pixel classification problem instead of measuring the similarity between images. Under this formulation, a novel few-shot semantic segmentation framework based on optimization meta-learning is proposed. Thanks to the proposed embedding module and linear model, our framework can achieve impressive performance. Furthermore, our method  provides a new alternative to deal with $K$-way, $N$-shot few-shot segmentation, which is more meaningful than the conventional setting.
%And the proposed method could be regarded as a baseline in this area.

\section{Acknowledge}
The work of Yang Gao is supported by Science and Technology Innovation 2030-``New Generation Artificial Intelligence'' Major Project (No. 2018AAA0100905), NSFC (No. 61432008)  and the Collaborative Innovation Center of Novel Software Technology and Industrialization. The work of Yinghuan Shi is supported by the Fundamental Research Funds for the Central Universities (No. 020214380056) and NSFC (No. 61673203).

\bibliographystyle{aaai}
\bibliography{ref}
\end{document}